\pdfoutput=1
\documentclass[letterpaper, 10 pt, conference]{ieeeconf}  

\IEEEoverridecommandlockouts                              

\usepackage{graphics} 
\usepackage{epsfig} 
\usepackage{mathptmx} 
\usepackage{times} 
\usepackage{amsmath} 
\usepackage{amssymb}  
\usepackage{multirow}
\usepackage{pifont}
\newcommand{\cmark}{\ding{51}}%
\newcommand{\xmark}{\ding{55}}%
\usepackage{graphicx}
\usepackage{subcaption}
\usepackage{multirow}
\usepackage[table,xcdraw]{xcolor}
\usepackage{bbold}
\usepackage{array}
\usepackage{booktabs}
\usepackage[linesnumbered,ruled]{algorithm2e}

\title{\LARGE \bf
UniBEVFusion: Unified Radar-Vision BEVFusion for 3D Object Detection
}

\author{\small
Haocheng Zhao,$^{1234\dagger}$,
Runwei Guan$^{1234\dagger}$,
Taoyu Wu$^{134}$,
Ka Lok Man$^{3}$,
Limin Yu$^{3\star}$,
Yutao Yue$^{514\star}$
\thanks{$^\star$Corresponding author.}
\thanks{\small$^{1}$ Institute of Deep Perception Technology, JITRI, Wuxi, China}
\thanks{\small$^{2}$ Department of Electrical Engineering and Electronics, University of Liverpool, Liverpool, UK}
\thanks{\small$^{3}$ Department of School of Advanced Technology, Xi'an Jiaotong-Liverpool University, Suzhou, China}
\thanks{\small$^{4}$ XJTLU-JITRI Academy of Technology, Xi'an Jiaotong-Liverpool University, Suzhou, China}
\thanks{\small$^{5}$ The Hong Kong University of Science and Technology (Guangzhou), Guangzhou, China}
}

\begin{document}

\maketitle
\thispagestyle{empty}
\pagestyle{empty}

\begin{abstract}
4D millimeter-wave (MMW) radar, which provides both height information and dense point cloud data over 3D MMW radar, has become increasingly popular in 3D object detection. In recent years, radar-vision fusion models have demonstrated performance close to that of LiDAR-based models, offering advantages in terms of lower hardware costs and better resilience in extreme conditions. However, many radar-vision fusion models treat radar as a sparse LiDAR, underutilizing radar-specific information. Additionally, these multi-modal networks are often sensitive to the failure of a single modality, particularly vision. To address these challenges, we propose the Radar Depth Lift-Splat-Shoot (RDL) module, which integrates radar-specific data into the depth prediction process, enhancing the quality of visual Bird’s-Eye View (BEV) features. We further introduce a Unified Feature Fusion (UFF) approach that extracts BEV features across different modalities using shared module. To assess the robustness of multi-modal models, we develop a novel Failure Test (FT) ablation experiment, which simulates vision modality failure by injecting Gaussian noise. We conduct extensive experiments on the View-of-Delft (VoD) and TJ4D datasets. The results demonstrate that our proposed Unified BEVFusion (UniBEVFusion) network significantly outperforms state-of-the-art models on the TJ4D dataset, with improvements of 1.44 in 3D and 1.72 in BEV object detection accuracy.

\end{abstract}

\section{INTRODUCTION}

Millimeter-wave (MMW) radar is widely used in roadside and vehicle-mounted transportation applications due to its reliable distance and velocity detection capabilities, even under extreme weather conditions \cite{wang2023multi,wang2023multi2,wang2024review}. However, the sparse nature of radar point cloud data and the lack of height information have posed challenges for accurate 3D object detection \cite{yao2023radar}. With recent advancements in 4D MMW radar technology, there is growing interest in utilizing this radar for 3D object detection, either as a standalone radar modality or fused with cameras \cite{zheng2023rcfusion,musiat2024radarpillars}. Radar-vision fusion has been shown to reduce hardware costs, enhance performance in extreme conditions, and maintain reasonable 3D object detection accuracy \cite{xiong2023lxl}.

In vision-based 3D object detection, a widely adopted approach is to project 2D image features into a Bird’s-Eye View (BEV) using intrinsic and extrinsic camera parameters along with accurate depth prediction \cite{philion2020lift,wang2023multi,wang2023multi2}. BEVFusion \cite{liu2023bevfusion}, a well-known LiDAR-Vision fusion model, provides an efficient architecture for fusing multi-modal data, improving upon methods like Lift-Splat-Shoot (LSS) \cite{philion2020lift} and pooling through optimizations and parallelization. Additionally, BEVFusion uses point cloud coordinates to assist with depth prediction, which is crucial for maintaining stability and accuracy in the model. Our reproduction shows competitive results in the radar-vision datasets, and our proposed UniBEVFusion network further improves the design.

However, in recent researches, radar has often been treated as a sparse LiDAR \cite{cui2023redformer}, and its specific characteristics are underutilized. A recent reproduction \cite{xiong2023lxl} of BEVFusion in radar-vision performs even similar to the results of pure radar detection. We argue that radar data should be fully leveraged in fusion models, and radar-specific information should be integrated into the depth prediction process to improve overall model performance. To address this, we propose Radar Depth LSS (RDL), which incorporates additional radar data, such as Radar Cross-Section (RCS), into the depth prediction process to enhance detection accuracy.

Moreover, multi-modal networks are particularly vulnerable to the failure of a single modality \cite{cheng2021robust,yao2023radar}, especially visual data. These networks often rely heavily on existence of both radar and image inputs, and their performance can degrade significantly when one modality is damaged or in adverse environment \cite{deng2022global,bansal2022radsegnet}. To evaluate the robustness of multi-modal models in such cases, we propose a novel ablation experiment called the Failure Test (FT), in which substantial noise is added to the visual input to simulate visual failure. As shown in our experiments, applying FT to BEVFusion results in a dramatic drop in performance, even below that of single-modal networks. To address this issue, we developed a novel multi-modal fusion module, Unified Feature Fusion (UFF), which unifies feature extraction and enhances features across different modalities to mitigate the impact of failure.

\begin{figure*}[ht]
\centering
\includegraphics[width=\linewidth]{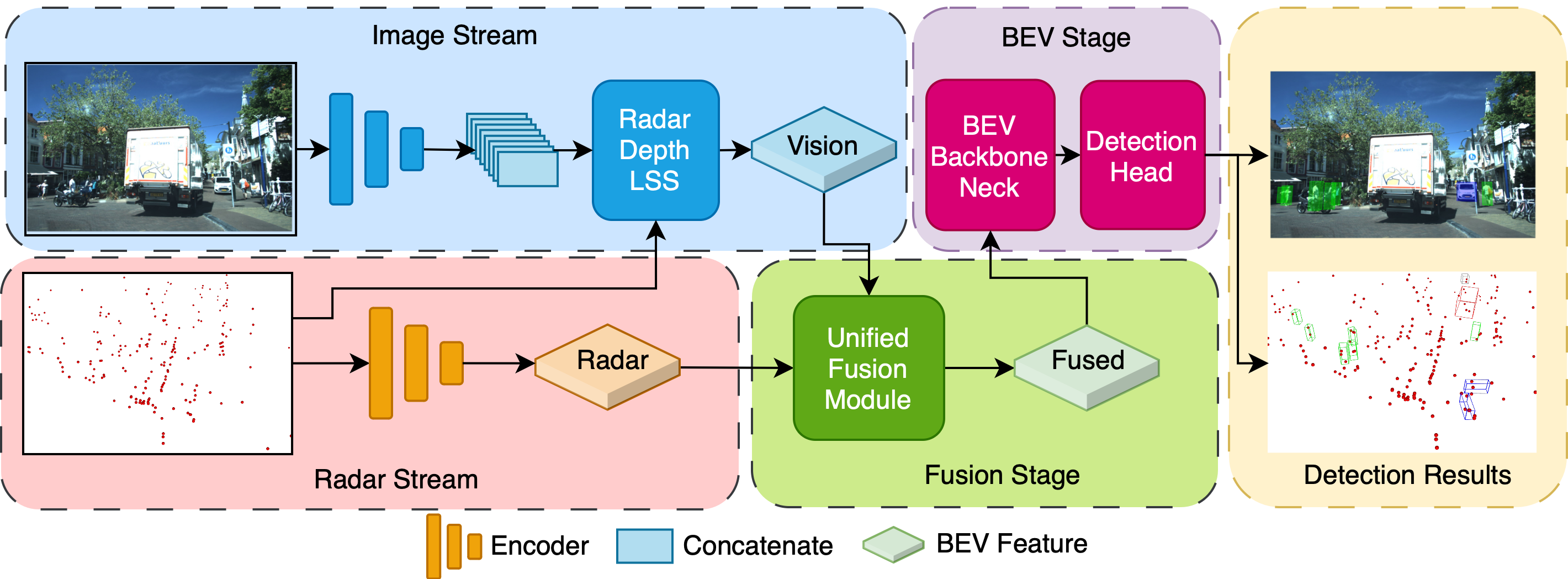}
\caption{Overview of the proposed UniBEVFusion network. The network consists of four main stages: Image, Radar, Fusion, and BEV. The Image and Radar stages are responsible for extracting BEV features from the image and radar, respectively. The Fusion stage is responsible for the fusion of the BEV features from the Image and Radar stages. The BEV stage is responsible for the final BEV feature extraction and 3D object detection head.}
\label{fig:overview}
\vspace{-0.4cm}
\end{figure*}

The contribution points of this paper are summarized as:
\begin{itemize}
        \item We propose the Radar Depth LSS (RDL) module, which integrates radar-specific information into the depth prediction process to improve the vision BEV feature transformation.
        \item We propose the novel fusion module Unified Feature Fusion (UFF) to extract features from different modalities and fuse them together. 
        \item We propose the novel Failure Test (FT) ablation experiment for multi-modal fusion in the case of near-failure of vision modality.
\end{itemize}

\section{Related Works}
\subsection{LiDAR Point Cloud 3D Object Detection}
Point cloud-based 3D object detection has evolved significantly with point-based, projection-based, and voxel-based methods \cite{wang2024review}. PointNet \cite{qi2017pointnet} and PointNet++ \cite{qi2017pointnet} capture global spatial information from raw point clouds but are computationally intensive due to their two-stage structure. Projection-based methods reduce computation cost by projecting point clouds into three 2D feature map \cite{li2016vehicle,yang2018pixor}. Voxel-based methods convert irregular point clouds into a regular voxel grid, reducing computational costs without sacrificing spatial feature resolution \cite{zhou2018voxelnet}. Building on voxel grids, PointPillars further optimize the computation by using pillars-based instead of voxels \cite{lang2019pointpillars}.

Point clouds provide accurate depth information, while images offer rich semantic information \cite{wang2023multi,wang2023multi2,contreras2024survey}. Aligning these modalities is fundamental to fusion networks. Camera data can be projected into the 3D coordinate system using intrinsic and extrinsic parameters, facilitating fusion with LiDAR point clouds \cite{wang2023multi,wang2023multi2,contreras2024survey}. To balance speed and performance, a common approach is to project both image and point cloud features into the bird’s-eye view (BEV) coordinate system. BEVFusion \cite{liu2023bevfusion} optimized the Lift-Splat-Shoot (LSS) pipeline and added point cloud projections to the camera coordinate system to aid in depth prediction \cite{philion2020lift}. Our proposed UniBEVFusion builds upon BEVFusion, optimizing radar feature integration for radar-vision fusion.

\subsection{Radar Point Cloud 3D Object Detection}
With the development of 4D millimeter-wave radar and the availability of open datasets, more researchers have explored radar-based object detection. Early work treated radar point clouds as a sparse LiDAR-like data \cite{cui2023redformer}, applying LiDAR object detection model such as PP-Radar \cite{apalffy2022}, reproduced BEVFusion \cite{xiong2023lxl}. Although promising, a significant gap remains between radar and LiDAR performance. Utilizing radar velocity \cite{musiat2024radarpillars}, radar coordination \cite{kim2023crn}, novel network modules \cite{broedermann2023hrfuser,palffy2022detecting, cui2023redformer,yu2023sparsefusion3d,peng2024mufasa}, LiDAR distillation \cite{zhao2024crkd}, adding gate \cite{fan2023grc,zhao2024crkd} and semantic alignment \cite{wu2023mvfusion} can improve the performance of radar-based object detection. 

In this paper, we focus on radar-vision feature fusion, which has shown promising results in recent studies. RADIANT \cite{long2023radiant} proposed a multi-stage fusion, including feature and detection head. FUTR3D \cite{chen2023futr3d} propose a modality-agnostic feature sampler to fuse radar, lidar, and camera. RCBEVDet \cite{lin2024rcbevdet} proposed an multi-head query-based method and a RCS-aware encoder that aligns BEV features using radar-specific information. RCFusion \cite{zheng2023rcfusion} generates pseudo-images from radar data and improves model performance with orthogonal feature transformations. LXL \cite{xiong2023lxl} enhances depth feature fusion by integrating radar and visual voxel features, achieving State-of-The-Art (SOTA) results on multiple radar-vision datasets. Our proposed UniBEVFusion will comparison the performance with these SOTA networks on the VoD \cite{apalffy2022} and TJ4D \cite{tj4d} datasets.


\section{Methodology}

\subsection{Overview}
Fig.\ref{fig:overview} shows the overall architecture of our proposed UniBEVFusion network, which contains four main parts: Image, Radar, Fusion, and BEV. Image and radar stream are responsible for extracting BEV features from the image and radar, respectively. The fusion stage handle the fusion of the BEV feature from the image and radar stream. The BEV stage is responsible for the final BEV feature extraction and 3D object detection head. 

Besides, the image encoder in image stream is a pre-trained swinTransformer \cite{liu2021swin}, which is used to extract features from the image. The radar encoder in radar stream and BEV stream is basically similar to PointPillar from the baseline of View-of-Delft (VoD) \cite{apalffy2022}, which use PillarFeatureNet, SECOND, and SECONDFPN \cite{yan2018second}. The 3D object detection head is a common 3D object detection head, which is used to predict the 3D bounding box and classification results.

\subsection{Radar Depth Lift-Splat-Shoot (RDL)}\label{sec:md_radar}

\begin{figure}[ht]
\centering
\includegraphics[width=\linewidth]{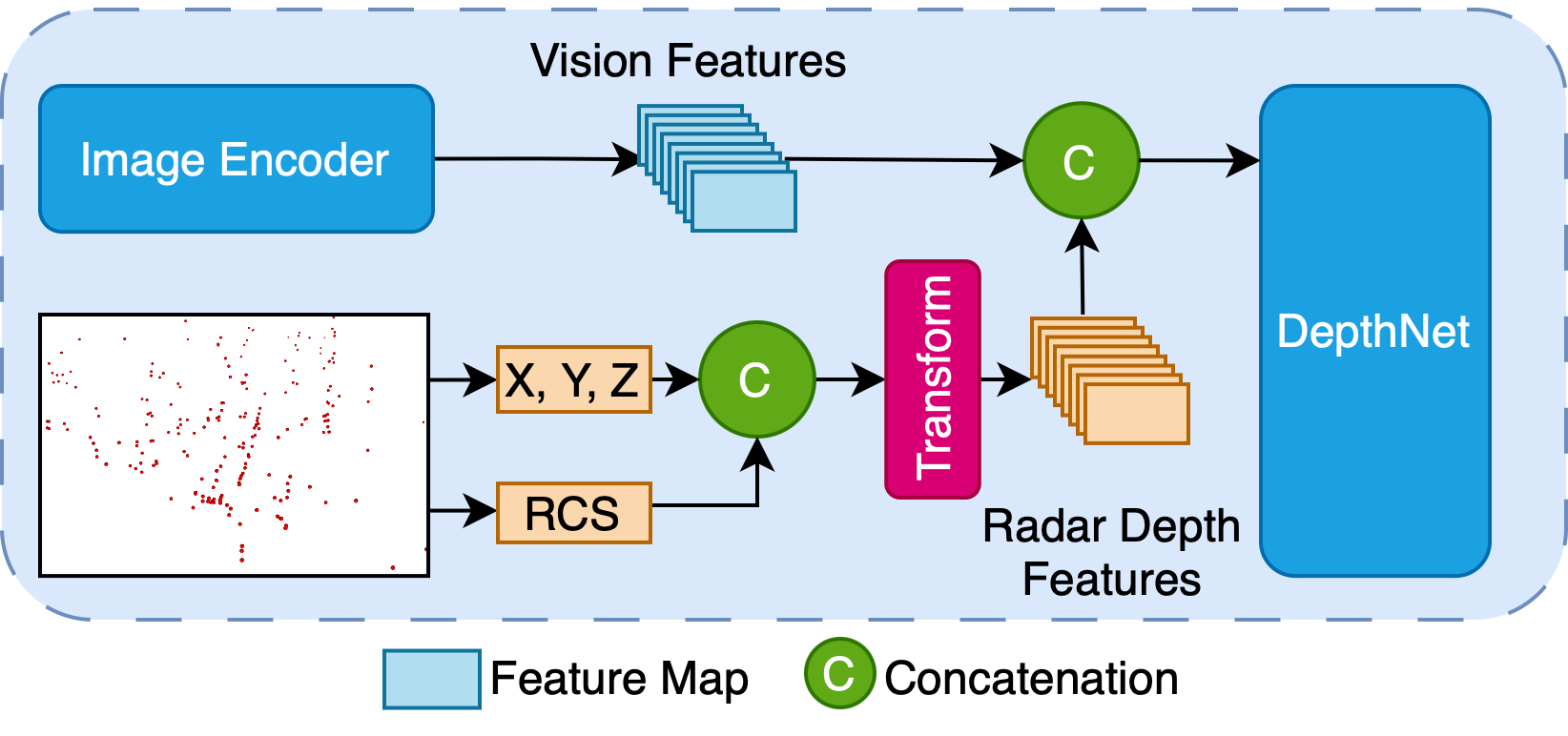}
\caption{Radar Depth Lift-Splat-Shoot (RDL) module.}
\label{fig:radar}
\end{figure}

LSS is an important milestone in visual-based 3D object detection \cite{philion2020lift}, but relies on correct depth prediction and computation is inefficient. BEVFusion \cite{liu2023bevfusion} provides a better optimzied LSS module, which gives projected point cloud as initial value of depth. Therefore, we inherit the View Transform module of BEVFusion and our RDL is based on this design. 

As shown in the Fig.\ref{fig:radar}, we first extract the coordination and RCS information, and then concat them to the depth prediction module. In fact, at this stage, we performes a early fusion of radar data and visual features. The extra information of point cloud data on VoD \cite{apalffy2022}, TJ4D \cite{tj4d}, and common LiDAR are shown in Table \ref{tab:radar}.
\begin{table}[ht]
\centering
\begin{tabular}{ll}
\hline
\rowcolor[HTML]{EFEFEF}
Sensor & Extra information \\
Radar in VoD & $x,\ y,\ z,\ RCS,\ V_r,\ V_r^\prime,\ t$ \\
Radar in TJ4D & $x,\ y,\ z,\ R,\ RCS,\ \alpha,\ \beta$ \\
LiDAR & $x,\ y,\ z,\ \text{intensity}$\\
\hline
\end{tabular}
\caption{Extra information of different sensors. $x, y, z$ are the coordinate information, $RCS$ is the radar signal strength, $V_r$ and $V_r^\prime$ are the relative and absolute velocity, $R$ is the distance, and $\alpha$ and $\beta$ are the horizontal and vertical angles.}
\label{tab:radar}
\end{table}

RCS is a key feature of radar data, which is related to the size, shape, and material of the object \cite{yao2023radar}. RDL reflects the physical characteristics of the objects in the depth prediction and retains this information in the later BEV features. The transform module is used to transform the radar depth features input channel number from $N+1$ to $64$, where $N$ and $1$ are the number of extra information channels (e.g., RCS, velocity) and depth information, respectively. 

\subsection{Unified Feature Fusion (UFF)} \label{sec:md_fusion}

\begin{figure}[ht]
\centering
\includegraphics[width=\linewidth]{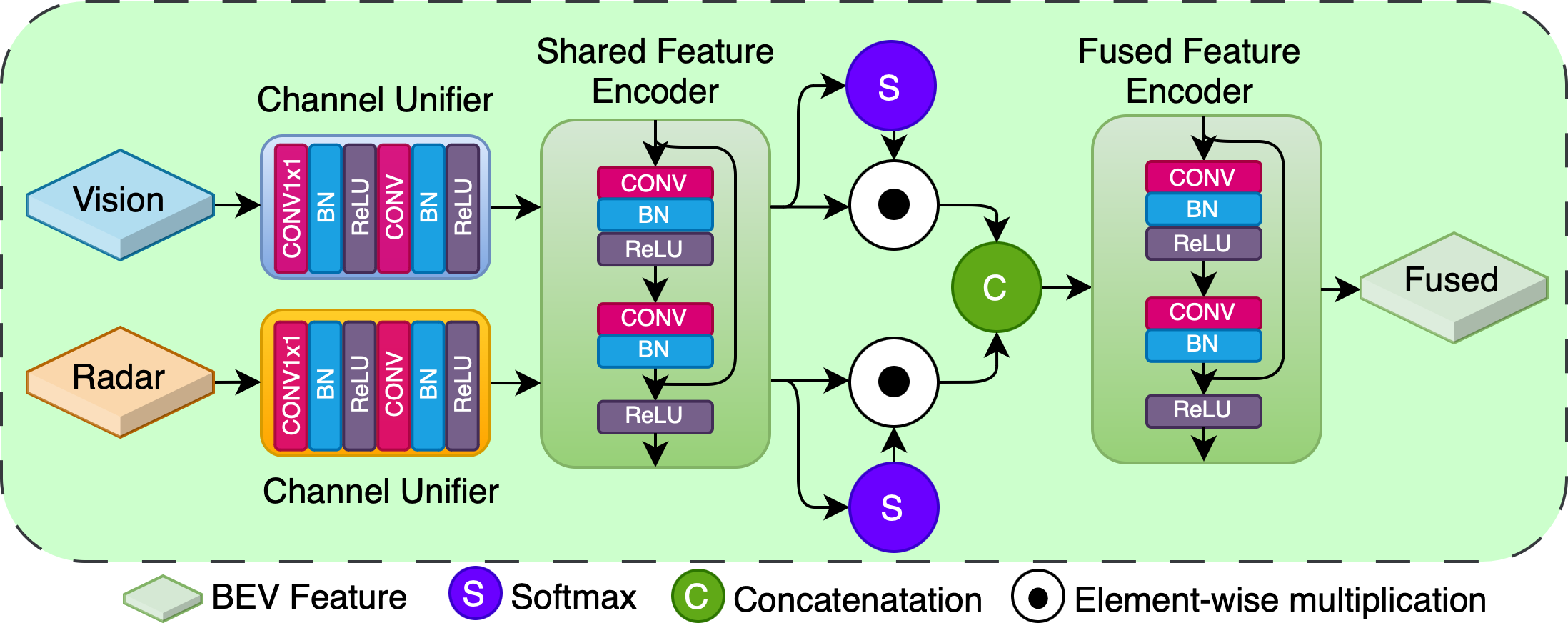}
\caption{Unified Feature Fusion (UFF).}
\label{fig:fusion}
\end{figure}

The UFF module, shown in Fig.\ref{fig:fusion}, is specifically designed to improve the reliability of multi-modal fusion by addressing the inherent differences between different sensor modalities. It consists of several key components: the Channel Unifier, the Shared Feature Encoder, the Softmax Concatenation Fusion, and the Fused Feature Encoder. The Channel Unifier aligns the feature dimensions of different modalities using 1x1 convolutions, ensuring a consistent channel representation across modalities. This not only simplifies the fusion process, but also enables more effective extraction of cross-modal features.

The Shared Feature Encoder plays a critical role in the normalization of feature representations from different modalities, mitigating discrepancies that may be due to modality-specific characteristics. Thus, it helps reduce performance degradation when a modality fails or provides suboptimal data. The softmax concatenation fusion integrates these processed features, while the use of softmax weighting allows the network to emphasize the most salient information across modalities, improving the overall quality of the feature fusion.

Both the Shared Feature Encoder and the Fused Feature Encoder are implemented as residual blocks, which facilitates deeper feature learning and promotes gradient flow during training. In addition to increasing the robustness of the fusion process, this architecture ensures that the fused features preserve essential information from each modality.




\subsection{Failure Test (FT)} \label{sec:md_ft}


In order to rigorously evaluate the robustness of the model under conditions where the vision modality fails, we propose a vision failure test. In contrast to the multi-view approach used in the CRN \cite{kim2023crn}, where robustness is evaluated for multiple views, we introduce Gaussian noise directly into the single-view image data sets to simulate the degradation of the visual input. This allows us to observe how detection performance changes with increasing noise level. The noisy image $I^\prime$ in FT$_\rho$ is defined as


\begin{equation}
\begin{aligned}
        I^\prime = I + \rho \cdot \mathbb{N}(0, \sigma^2),
\end{aligned}
\end{equation}


\begin{table*}[ht]
\centering
\begin{tabular}{c|ccc|c|ccc|c}
\multirow{2}{*}{\textbf{Model}} &
        \multicolumn{4}{c|}{\textbf{Entrire Annotated Area}} &
        \multicolumn{4}{c}{\textbf{Driving Corridor Area}} \\ \cmidrule{2-9}
        &
        \multicolumn{1}{c}{\textbf{Car}} &
        \multicolumn{1}{c}{\textbf{Ped}} &
        \multicolumn{1}{c}{\textbf{Cyc}} &
        \multicolumn{1}{|c|}{\textbf{mAP}} &
        \multicolumn{1}{c}{\textbf{Car}} &
        \multicolumn{1}{c}{\textbf{Ped}} &
        \multicolumn{1}{c}{\textbf{Cyc}} &
        \multicolumn{1}{|c}{\textbf{mAP}} \\ \midrule
\textbf{BEVFusion$^*$}     & 42.02 & 38.98 & 67.54 & 49.51 & 72.23 & 48.67 & 85.57 & 69.02 \\
\textbf{RCFusion}       & 41.70 & 38.95 & 68.31 & 49.65 & 71.87 & 47.50 & 88.33 & 69.23 \\
\textbf{FUTR3D}         & 46.01 & 35.11 & 65.98 & 49.03 & 78.66 & 43.10 & 86.19 & 69.32 \\
\textbf{GRC-Net}        & 27.90 & 31.00 & 64.60 & 41.10 &   -   &   -   &   -   &   -   \\
\textbf{RCBEVDet}          & 40.60 & 38.80 & 70.40 & 49.90 & 72.40 & 49.80 & 87.00 & 69.80 \\ 
\textbf{LXL}            & 42.33 & 49.48 & 77.12 & 56.31 & 72.18 & 58.30 & 88.31 & 72.93 \\ \midrule
\textbf{BEVFusion}      & 40.85 & 47.60 & 72.92 & 53.79 & 71.93 & 57.10 & 88.23 & 72.42 \\
\textbf{UniBEVFusion} & 42.22 & 47.11 & 72.94 & 54.09 & 72.10 & 57.71 & 93.29 & 74.37 \\
\bottomrule
\end{tabular}
\caption{Results on VoD. BEVFusion$^*$ is the reproduction results from LXL \cite{xiong2023lxl}.}
\label{tab:vod}
\vspace{-0.4cm}
\end{table*}    

where $\rho$ is the noise level, $I$ is the original clean image, $I^\prime$ is the noise corrupted image, and $\mathbb{N}(0, \sigma^2)$ is Gaussian noise with mean 0 and variance $\sigma^2$. By systematically varying $\rho$, we evaluate the performance of the model under different noise intensities. As presented in section \ref{sec:exp_ft}, both BEVFusion and our proposed UniBEVFusion show sensitivity to noise in the visual modality, highlighting the impact of modality-specific degradation on overall model performance. This analysis underscores the importance of robust multi-modal fusion in maintaining detection accuracy even under adverse conditions.

\section{Experiments}
We first give the brief introduction to the datasets used in the experiments in Section \ref{sec:exp_data}. Then we compare the performance of different models on VoD \cite{apalffy2022} and TJ4D \cite{tj4d} in Section \ref{sec:exp_vod} and Section \ref{sec:exp_tj4d}, respectively. The results of FT and the ablation study is shown in Section \ref{sec:exp_ft}. Lastly, we test the performance of different image resolutions in Section \ref{sec:exp_res}.

\subsection{Datasets} \label{sec:exp_data}
The datasets used in this paper, VoD \cite{apalffy2022} and TJ4D \cite{tj4d}, both provide 4D MMW radar data. Radar point clouds in VoD includes [$x,\ y,\ z,\ RCS,\ V_r,\ V_r^\prime,\ t$], while TJ4D includes [$x,\ y,\ z,\ R,\ RCS,\ \alpha,\ \beta$], where $x, y, z$ represent coordinates, $RCS$ is radar signal strength, $V_r$ and $V_r^\prime$ are relative and absolute velocities, $R$ is distance, and $\alpha$, $\beta$ are angles.

The VoD dataset includes categories for car, pedestrian, and cyclist, while TJ4D adds trucks. We followed the official method, segmenting VoD’s 6435 frames into 5139 for training and 1296 for validation, and TJ4D’s 7757 frames into 5717 for training and 2040 for validation. In this paper, our experimental camera resolutions are resized to [608, 968] for VoD and [480, 640] for TJ4D.

For evaluation, we used Mean Average Precision (mAP) with IoU thresholds of 0.5 for cars/trucks and 0.25 for pedestrians/bicycles. VoD’s official evaluation includes RoI 3D detection within [$-4\le x \le 4m, z\le 25m$], while TJ4D evaluates 3D and BEV detection across all ranges.



\subsection{Results on VoD} \label{sec:exp_vod}
Table \ref{tab:vod} shows the performance of our model on the validation set of VoD \cite{apalffy2022}, where the $mAP$ is slightly lower than that of the LXL fusion network in the Entire Annotation Area (EAA). LXL achieves State-of-the-Art (SOTA) performance across the multi-modal radar datasets. However, in the more critical Driving Corridor Area (DCA), which is constrained by distance, UniBEVFusion outperforms LXL. While our model performs slightly worse than LXL in the detection of cars and pedestrians, it significantly outperforms LXL in the detection of cyclists. Overall, UniBEVFusion shows superior performance in the DCA, which is crucial for autonomous driving tasks, and maintains competitive results in the EAA, where it outperforms the other algorithms.

Furthermore, it is noteworthy that our reproduced BEVFusion outperforms previously reported results \cite{xiong2023lxl}. By modifying the detection head and radar PillarFeatureNet to align with UniBEVFusion, we have achieved a higher level of performance. This improved BEVFusion serves as a robust baseline for evaluating the effectiveness of our proposed UniBEVFusion network.



\begin{figure*}[ht]
\centering
\includegraphics[width=0.98\linewidth]{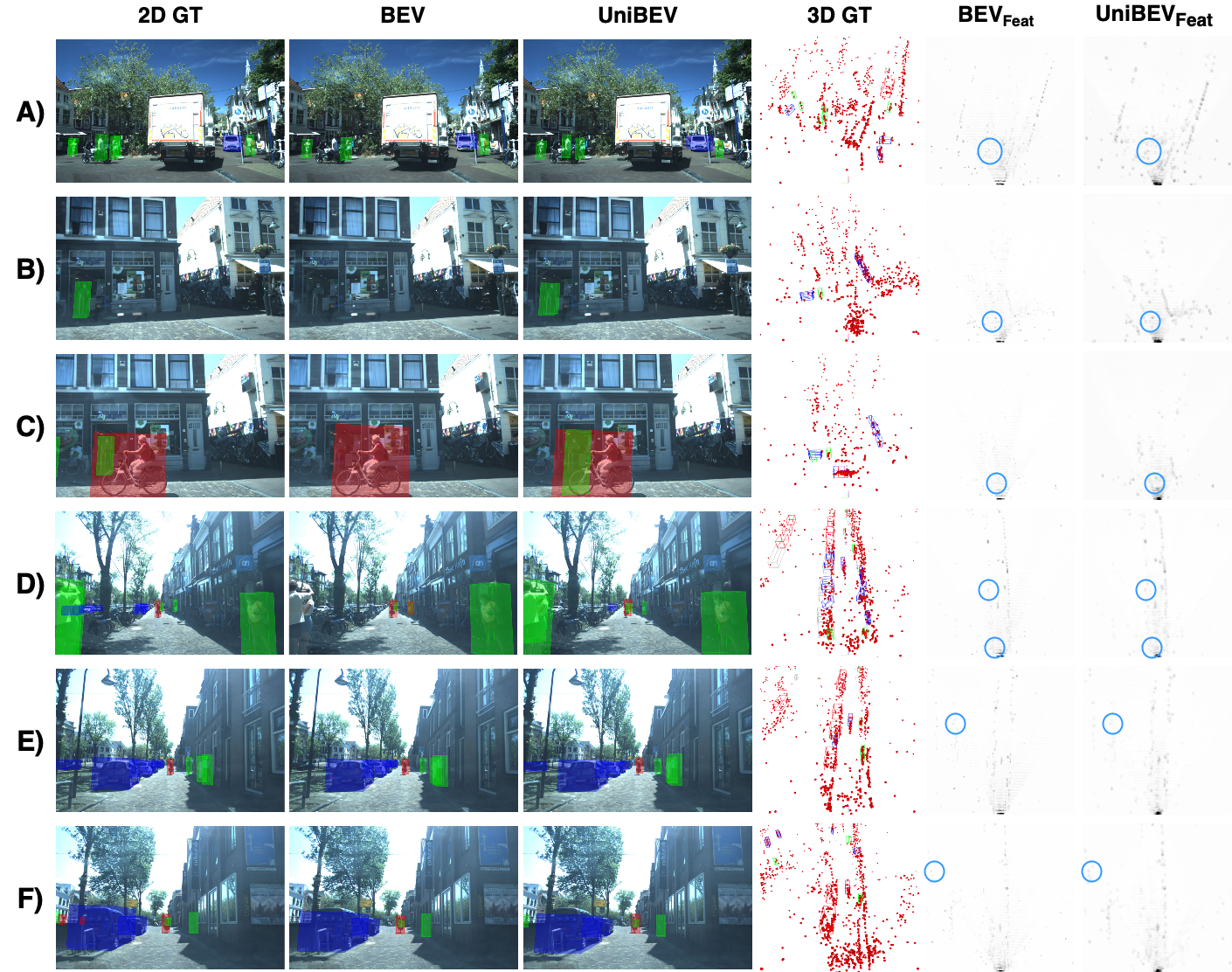}
\caption{Comparison of detection results between UniBEVFusion and BEVFusion \cite{liu2023bevfusion}. 2D GT and 3D GT are the ground truth of 2D and 3D detection, respectively. The BEV and BEV$_{\text{Feat}}$ are the detection results and fused BEV feature of BEVFusion, respectively. The UniBEV and UniBEV$_{\text{Feat}}$ are the detection results and fused BEV feature of UniBEVFusion, respectively. Red, green, and blue boxes represent cars, pedestrians, and cyclists, respectively.}
\label{fig:vod5f}
\vspace{-0.4cm}
\end{figure*}
Results in Fig.\ref{fig:vod5f} validate the performance of UniBEVFusion compared to Ground Truth (GT) and BEVFusion \cite{liu2023bevfusion}. The right section of the figure shows the fused BEV features, where UniBEVFusion covers a larger area than BEVFusion, though with lower overall feature magnitudes due to the Softmax layer in the UFF module. Despite this, the features in key regions remain strong, and the UFF module effectively extracts features from different modalities, providing broader context and more stable fused features for object detection.

UniBEVFusion demonstrates superior performance in handling occlusions (Fig.\ref{fig:vod5f} A, C, E, F), where its larger feature field allows it to detect occluded objects more reliably, reducing the likelihood of dismissing them as noise. In shadowed and partially occluded scenarios (Fig.\ref{fig:vod5f} B, C), where vision alone struggles, UniBEVFusion accurately identifies the target using radar-specific information from the RDL module. Additionally, in close-range detection (Fig.\ref{fig:vod5f} D), UniBEVFusion succeeds where BEVFusion fails, likely due to the latter’s lack of sufficient contextual information in the fused BEV feature. Overall, UniBEVFusion performs better in occlusion, shadow, and both short- and long-range detection, with the UFF and RDL modules enhancing performance in various scenarios.

\subsection{Results on TJ4D} \label{sec:exp_tj4d}
Compared to the VoD dataset’s point cloud range [[0, 51.2], [-25.6, 25.6], [-3, 2]] \cite{apalffy2022}, the TJ4D dataset covers a significantly larger range [[0, 69.12], [-39.68, 39.68], [-4, 2]] \cite{tj4d}, which introduces additional complexity for 3D object detection. Despite this increased difficulty, the performance of UniBEVFusion on TJ4D, as shown in Table \ref{tab:tj4d}, is consistent with its results on VoD, and it even surpasses the validation outcomes of the LXL algorithm \cite{xiong2023lxl}.

UniBEVFusion achieves improvements of 1.44 and 1.72 over LXL in 3D object detection and BEV accuracy, respectively. Notably, in the Car detection task, it outperforms RCFusion \cite{zheng2023rcfusion} by 5.54 in 3D detection and by 9.37 in BEV detection. These results highlight the effectiveness of the RDL and UFF modules, which significantly enhance the model’s performance and robustness, making UniBEVFusion particularly well-suited for 3D object detection in more challenging and expansive environments.



\begin{table*}[ht]
\centering
\begin{tabular}{c|cccc|c|cccc|c}
\multirow{2}{*}{\textbf{Model}} & \multicolumn{5}{c|}{\textbf{3D}}       & \multicolumn{5}{c}{\textbf{BEV}}      \\ \cmidrule{2-11}
        &
        \multicolumn{1}{c}{\textbf{Car}} &
        \multicolumn{1}{c}{\textbf{Ped}} &
        \multicolumn{1}{c}{\textbf{Cyc}} &
        \multicolumn{1}{c}{\textbf{Tru}} &
        \multicolumn{1}{|c|}{\textbf{mAP}} &
        \multicolumn{1}{c}{\textbf{Car}} &
        \multicolumn{1}{c}{\textbf{Ped}} &
        \multicolumn{1}{c}{\textbf{Cyc}} &
        \multicolumn{1}{c}{\textbf{Tru}} &
        \multicolumn{1}{|c}{\textbf{mAP}} \\ \midrule
\textbf{MVX-Net}                & 22.28 & 19.57 & 50.70 & 11.21 & 25.94 & 37.46 & 22.70 & 54.69 & 18.07 & 33.23 \\
\textbf{FUTR3D}                 &   -   &   -   &   -   &   -   & 32.42 &   -   &   -   &   -   &   -   & 37.51 \\
\textbf{RCFusion}               & 29.72 & 27.17 & 54.93 & 23.56 & 33.85 & 40.89 & 30.95 & 58.30 & 28.92 & 39.76 \\
\textbf{LXL}                    &   -   &   -   &   -   &   -   & 36.32 &   -   &   -   &   -   &   -   & 41.20 \\ \midrule
\textbf{BEVFusion}              & 38.09 & 29.45 & 51.26 & 23.73 & 35.63 & 48.53 & 32.04 & 55.40 & 28.96 & 41.23 \\ 
\textbf{UniBEVFusion}           & 44.26 & 27.92 & 51.11 & 27.75 & 37.76 & 50.43 & 29.57 & 56.48 & 35.22 & 42.92 \\\bottomrule
\end{tabular}
\caption{Comparison of the results on TJ4D.}
\label{tab:tj4d}
\vspace{-0.1cm}
\end{table*}

\subsection{Failure Test} \label{sec:exp_ft}

\begin{table*}[ht]
\centering
\begin{tabular}{c|cc|c|ccc|c|ccc}
\multirow{2}{*}{\textbf{Model}} & \multirow{2}{*}{\textbf{RDL}} & \multirow{2}{*}{\textbf{UFF}} & \multicolumn{4}{c}{\textbf{TJ4D 3D mAP}}  & \multicolumn{4}{|c}{\textbf{TJ4D BEV mAP}} \\ \cmidrule{4-11}
                        &  &  & \textbf{FT$_0$} & \textbf{FT$_{0.5}$} & \textbf{FT$_{0.7}$} & \textbf{FT$_{0.9}$} & \textbf{FT$_0$} & \textbf{FT$_{0.5}$} & \textbf{FT$_{0.7}$} & \textbf{FT$_{0.9}$} \\ \midrule
\textbf{BEVFusion}      & \xmark  & \xmark  & 35.63        & 21.03          & 17.17          & 11.43          & 41.23        & 26.49          & 21.06          & 14.03          \\
\textbf{BEVFusion}  & \cmark  & \xmark  & 36.23        & 21.23          & 17.26          & 11.84          & 41.98        & 26.80          & 21.71          & 14.53          \\
\textbf{UniBEVFusion}     & \xmark  & \cmark  & 36.84        & 22.54          & 17.61          & 12.01          & 42.49        & 27.19          & 21.81          & 15.55          \\
\textbf{UniBEVFusion}   & \cmark  & \cmark  & 37.76        & 22.79          & 17.44          & 12.47          & 42.92        & 27.70          & 22.11          & 16.17          \\ \midrule
\multirow{2}{*}{\textbf{Model}} & \multirow{2}{*}{\textbf{RDL}} & \multirow{2}{*}{\textbf{UFF}}          & \multicolumn{4}{c}{\textbf{VoD ALL}} & \multicolumn{4}{|c}{\textbf{VoD RoI}} \\ \cmidrule{4-11}
                        &  &  & \textbf{FT$_0$} & \textbf{FT$_{0.5}$} & \textbf{FT$_{0.7}$} & \textbf{FT$_{0.9}$} & \textbf{FT$_0$} & \textbf{FT$_{0.5}$} & \textbf{FT$_{0.7}$} & \textbf{FT$_{0.9}$} \\ \midrule
\textbf{BEVFusion}      & \xmark  & \xmark  & 53.79        & 41.04          & 36.78          & 30.37          & 72.42        & 56.13          & 50.01          & 44.32          \\
\textbf{BEVFusion}  & \cmark  & \xmark  & 53.77        & 41.15          & 36.80          & 30.35          & 74.02        & 56.24          & 50.69          & 44.53          \\
\textbf{UniBEVFusion}     & \xmark  & \cmark  & 53.70        & 41.42          & 37.69          & 31.70          & 72.50        & 58.00          & 51.04          & 44.27          \\
\textbf{UniBEVFusion} & \cmark  & \cmark  & 54.09        & 41.69          & 37.26          & 33.03          & 74.37        & 58.87          & 51.74          & 45.33          \\ \bottomrule
\end{tabular}
\caption{Comparison between BEVFusion \cite{liu2023bevfusion} and UniBEVFusion in Failure Test (FT).}
\label{tab:ft}
\vspace{-0.2cm}
\end{table*}

Based on the previous introduction, we evaluate BEVFusion \cite{liu2023bevfusion} and our proposed UniBEVFusion model using $\rho = [0.5, 0.7, 0.9]$. Since the design of the noise is related to the random numbers, the average of 10 operations was taken for all the test results. Results in Table \ref{tab:ft} shows the performance of baseline FT$_0$, and evaluations FT$_{0.5}$, FT$_{0.7}$, and FT$_{0.9}$. As the baseline FT$_0$ is the normal evaluation results of these model, thus, we will also discuss the effectiveness of the RDL and UFF in this section.

For BEVFusion model in TJ4D FT evaluation, adding RDL improves much in baseline performance, but the FT results are close. RDL is designed for accurate depth prediction in image stream, and it can not guarantee the robust results when image is failure. Adding UFF improves both the baseline and FT performance, which indicates that the UFF is effective in improving the robustness of the model. As for the UniBEVFusion model, the conclusion is similar to the BEVFusion model, and the overall FT results are better than BEVFusion. 

In VoD FT evaluation, conclusion are different in two different evaluation range. For entire annotation area, the results are close, and we can not tell the effectiveness of the RDL on this dataset. The UFF bring much leading than the BEVFusion model, which indicates that the UFF is effective in improving the robustness of the model. However, for the driving corridor area, the basic conclusion are similar to the TJ4D. Moreover, with the noise level $\rho$ increasing, the performance gap between the two models is getting smaller for our UniBEVFusion in all evaluation.

On top of the results, we can conclude that the UFF and RDL are effective in improving the performance of multi-modal model. Besides, UFF provides a better robustness in the case of vision failure.

\subsection{Image Resolution} \label{sec:exp_res}

\begin{table}[ht]
\begin{tabular}{c|c|c|rr|rr}
\multicolumn{1}{c|}{\textbf{Scale}} &
        \textbf{Image Size} &
        \textbf{RDL} &
        \multicolumn{1}{c}{\textbf{3D}} &
        \multicolumn{1}{c}{\textbf{$\Delta$ (\%)}} &
        \multicolumn{1}{|c}{\textbf{BEV}} &
        \multicolumn{1}{c}{\textbf{$\Delta$ (\%)}} \\ \midrule
\textbf{1.00} & \textbf{{[}960, 1280{]}} & \xmark & 12.02 & 0.0\%   & 14.74 & 0.0\%   \\
\textbf{1.00} & \textbf{{[}960, 1280{]}} & \cmark & 13.19 & 9.8\%   & 26.73 & 5.9\%  \\ \midrule
\textbf{0.75} & \textbf{{[}720, 960{]}}  & \xmark & 14.81 & 0.0\%   & 17.85 & 0.0\%   \\
\textbf{0.75} & \textbf{{[}720, 960{]}}  & \cmark & 16.91 & 14.2\%  & 30.39 & 3.9\%  \\ \midrule
\textbf{0.50} & \textbf{{[}480, 640{]}}  & \xmark & 13.66 & 0.0\%   & 16.72 & 0.0\%   \\
\textbf{0.50} & \textbf{{[}480, 640{]}}  & \cmark & 14.46 & 5.8\%   & 29.68 & 4.3\%  \\ \midrule
\textbf{0.25} & \textbf{{[}240, 320{]}}  & \xmark & 7.54  & 0.0\%   & 7.63  & 0.0\%   \\
\textbf{0.25} & \textbf{{[}240, 320{]}}  & \cmark & 6.44  & -14.6\% & 10.02 & -2.1\% \\ \bottomrule
\end{tabular}
\caption{Comparison of different image resolutions on TJ4D.}
\label{tab:res}
\vspace{-0.1cm}
\end{table}

In RCFusion, they shows that larger image sizes have a positive impact on the fusion model results, but also increase the arithmetic consumption and decrease the FPS. Immediately following the discussion on image sizes, we test the performance of the pure camera modality on the TJ4D \cite{tj4d} dataset for different image scaling as well as validate our proposed RDL. It is worth noting that although we are testing the performance of pure image data, BEVFusion still uses coordination information from the point cloud to assist depth prediction.


Evaluating scaling from 0.25 to 0.75 shows a consistent trend with RCFusion, where smaller scales result in missing information and reduced performance. Interestingly, full-size images performed worse than 0.5 and 0.75 due to the model being tuned for 0.5 and overfitting on detailed images. The 0.25 scale yielded the worst results due to excessive detail loss and sparse features after BEV transformation. Despite the slightly better performance at 0.75, we opted for 0.50 scaling for operational speed.


Moreover, comparing the effectiveness of our proposed RDL, the results of 0.50~1 outperform the original BEVFusion \cite{liu2023bevfusion} by at least 5.84\% and 3.88\% in 3D and BEV, respectively. However, the performance at 0.25 is reduced by 14.6\% and 2.1\%, respectively. In the absence of image information, image features and coordinate information are misaligned. RCS information representing the shape, material, and size of the object is also incorrectly added to features, resulting in learning wrong features and worse performance.

\section{Conclusion}
In this paper, we demonstrated that the UniBEVFusion network achieves state-of-the-art performance on the TJ4D \cite{tj4d} and driving corridor of the View-of-Delft (VoD) datasets \cite{apalffy2022}. The results indicate that UniBEVFusion significantly improves detection performance, particularly in challenging conditions such as shadows, occlusions, short-range, and long-range scenarios. Our proposed Radar Depth Lift-Splat-Shoot (RDL) module and Unified Feature Fusion (UFF) framework are effective in enhancing the model’s performance. Specifically, RDL integrates radar depth and RCS information into the depth prediction process, boosting the accuracy of vision-based 3D object detection. UFF mitigates the model’s reliance on the simultaneous availability of multiple modalities, improving its robustness against single-modality failures. Although Gaussian noise was the only simulation solution used in the Failure Test (FT), it still provided valuable insights into the model’s robustness. In future work, we plan to further optimize UFF and RDL to improve the performance of multi-modal models in scenarios where one modality fails. In addition, we will incorporate more diverse failure modes into the FT and develop more precise evaluation metrics to better assess robustness.

\newpage
\bibliographystyle{IEEEbib}
\bibliography{ref}

\end{document}